\documentclass[11pt]{article}
 
\pdfoutput=1

\usepackage{acl}

\usepackage{times}
\usepackage{latexsym}

\usepackage{arydshln} 
\usepackage{makecell} 
\usepackage{booktabs} 

\usepackage{amsfonts}
\usepackage{amssymb} 
\usepackage{arydshln}
\usepackage[T1]{fontenc}

\usepackage[utf8]{inputenc}

\usepackage{microtype}

\usepackage{inconsolata}

\usepackage{graphicx}
\usepackage{subcaption}
\expandafter\def\csname ver@subfig.sty\endcsname{}
\usepackage{float}
\usepackage{amsmath}

\usepackage{subfig}
\usepackage{multirow}
\usepackage{makecell} 

\title{TEGEE: Task dEfinition Guided Expert Ensembling for Generalizable and Few-shot Learning}

\author{
    \thanks{Equal contribution}Xingwei Qu\textsuperscript{2,3}\quad
    \footnotemark[1]Yiming Liang\textsuperscript{4,5,\bf10}\quad
    \textbf{Yucheng Wang}\textsuperscript{1}\quad
    \textbf{Tianyu Zheng}\textsuperscript{1,9}\\
    \textbf{Tommy Yue}\textsuperscript{7}\quad
    \textbf{Xingyuan Bu}\textsuperscript{1}\quad
    \textbf{Lei Ma}\textsuperscript{\bf8,\bf10}\quad
    \textbf{Stephen W. Huang}\textsuperscript{\bf11}\quad
    \textbf{Jiajun Zhang}\textsuperscript{\bf4,5}\\
    \textbf{Yinan Shi}\textsuperscript{\bf1}\quad
    \thanks{Corresponding author.}\textbf{Chenghua Lin}\textsuperscript{\bf3}\quad
    \footnotemark[2]\textbf{Jie Fu}\textsuperscript{\bf2}\quad
    \footnotemark[1]\footnotemark[2]\textbf{Ge Zhang}\textsuperscript{\bf1,2,6}\\
    \textsuperscript{1}M-A-P\quad
    \textsuperscript{2}HKUST \quad 
    \textsuperscript{3} University of Manchester\quad
    \textsuperscript{4}Institute of Automation, CAS\\
     \textsuperscript{5}University of Chinese Academy of Sciences\quad
    \textsuperscript{6}University of Waterloo\quad
    \textsuperscript{7}Ohio State University\\
    \textsuperscript{8}Peking University\quad
    \textsuperscript{9}Beijing University of Posts and Telecommunications\quad
    \textsuperscript{10}BAAI\quad
    \textsuperscript{11}Harmony.ai\\
    \vspace{-4ex}
\small
\texttt{ge.zhang@uwaterloo.ca, jiefu@ust.hk, chenghua.lin@manchester.ac.uk} \\
}

\begin{document}
\maketitle
\begin{abstract}

Large Language Models (LLMs) exhibit the ability to perform in-context learning (ICL), where they acquire new tasks directly from examples provided in demonstrations. This process is thought to operate through an implicit task selection mechanism that involves extracting and processing task definitions from these demonstrations. However, critical questions remain: Which is more essential---task extraction or definition? And how can these capabilities be further improved? To address these questions, we propose \textbf{TEGEE} (Task Definition Guided Expert Ensembling), a method that explicitly extracts task definitions and generates responses based on specific tasks. Our framework employs a dual 3B model approach, with each model assigned a distinct role: one focuses on task definition extraction, while the other handles learning from demonstrations. This modular approach supports the hypothesis that extracting task definitions is more vital than processing the task itself. Empirical evaluations show that TEGEE performs comparably to the larger LLaMA2-13B model. By leveraging a modular design, our approach extends traditional ICL from few-shot to many-shot learning, supporting an unlimited number of demonstrations and enhancing continual learning capabilities.

\end{abstract}

\section{Introduction}

\begin{figure*}[htb]
    \centering
    \includegraphics[width=0.9\textwidth]{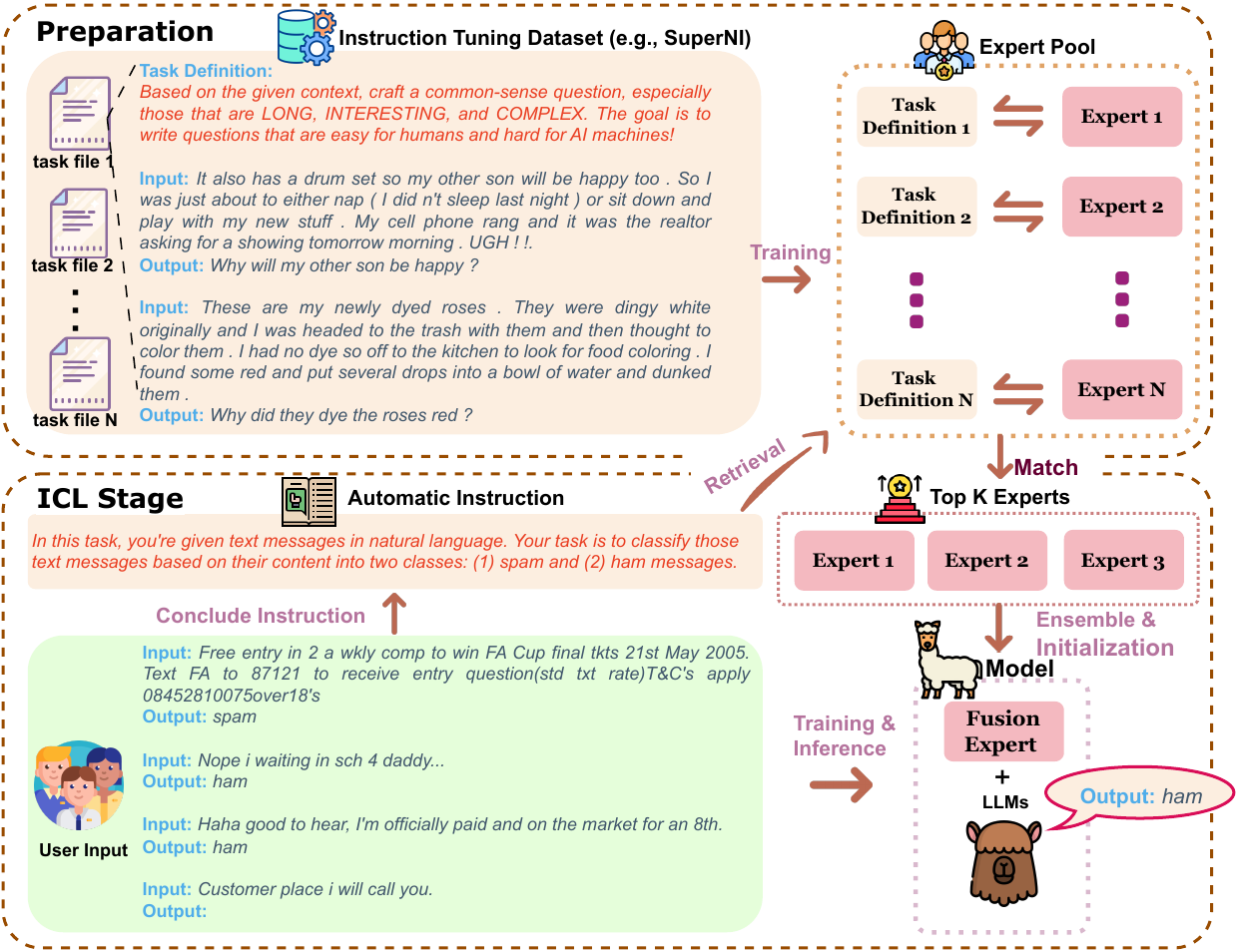}
    \caption{Pipeline of TEGEE. The preparation stage focuses on generating a pool of task-based experts, each trained using task-specific data. 
    ICL firstly generates task definitions from given demonstrations using a task definition generator, which can be a custom generator, GPT, or human. 
    Secondly, we utilize the task definitions to retrieve the task-based expert pool and match the top three experts that are most similar.
    Next, the expert ensemble step initializes the fusion expert by performing a weighted average of the matched experts' weights. Finally, the data format is reconstructed for training the fusion expert and performing inference on the last input field.}
    \label{fig:pipline}
\end{figure*}

In recent years, large language models (\textbf{LLMs}) have showcased remarkable capabilities, particularly for in-context learning (\textbf{ICL}). 
This method allows LLMs to learn how to execute specific tasks by analyzing a few labeled examples provided in their input context, without any updates to their parameters. 
This breakthrough enables LLMs to serve as versatile, general-purpose models capable of tackling a wide range of tasks with minimal guidance, as highlighted by \citet{brown2020language}. However, there is still no consensus on when or how ICL works. 

However, ICL also has its disadvantages. 
Most open-sourced LLMs such as LLaMA \cite{touvron2023llama}, Falcon \cite{falcon40b}, and ChatGLM \cite{du2022glm,zeng2022glm} have fixed input lengths, limiting the number of demonstrations that can be provided. 
Even with sufficient demonstrations, the performance of ICL cannot be enhanced. This limitation is due to ICL's selective nature, requiring the choice of representative demonstrations, as proven by \citet{levy2023diverse}. 


In this paper, we propose a novel ICL-style ensembling approach through the lens of few-shot learning, called \textbf{T}ask d\textbf{E}finition \textbf{Guided} \textbf{E}xpert \textbf{E}nsembling (\textbf{TEGEE}). 
Inspired by \citet{pan2023incontext}, our approach incorporates two 3B models: one focuses on task definition determination, and the other specializes in learning from task-specific demonstrations. 
Since few-shot learning usually targets specific tasks, it is difficult to generalize to other unseen tasks.  
To overcome this problem, we create a dynamic expert pool filled with prior knowledge during the demonstration learning phase.
In other words, the experts in the pool can be selected according to the task definition derived in the task definition generation phase. 
Furthermore, this dynamic pool utilizes LoRA \citep{hu2021lora} for selecting the top experts based on task requirements. 
Our experiments show that even with continual training on just five samples, our ensembling methods outperform both traditional ICL and non-ensembling approaches. 
This setup supports lifelong learning by surpassing the fixed limitations of LLMs, significantly improving performance. 

We evaluate TEGEE on 117 subtasks in the official Super Natrual Instruction (\textbf{SuperNI}) \cite{supernaturalinstructions} test dataset split \cite{supernaturalinstructions}. Our experiments, comparing the performance to LLaMA2 models with 7B and 13B parameters \cite{touvron2023llama}, and Falcon 7B \cite{falcon40b}, show that our models significantly outperform these two 7B level models and exhibit performance that is marginally below the LLaMA2-13B model, indicating our approach is competitive within a narrow margin. This result demonstrates that our solution is effective in addressing the limitations of ICL and few-shot learning techniques.

In conclusion, our contributions are as follows:
\begin{itemize}
    \item We propose a novel framework that delineates the roles between two models: \textbf{task definition} and \textbf{task processing}. This distinction further confirms that the primary challenge of in-context learning (ICL) is the \textbf{extraction of task definitions, rather than the processing of tasks itself}.
    \item Through extensive experimentation, we have validated that the synergistic efficacy of our dual 3B model approach is on par with the ICL performance of the LLaMA2-13B model. 
    \item Our framework's expert pool is designed to dynamically augment its database with input demonstrations, thereby enhancing and enriching its capacity on a continual basis. 
\end{itemize}

\section{Related Work}

\subsection{ICL and Few-Shot Learning with LLMs}

The emergence of LLMs has revolutionized the landscape of machine learning, particularly for ICL and few-shot learning~\citep{brown2020language}. 
These methodologies have been instrumental in enabling rapid model adaptation to new tasks, even with limited examples. 
This section delves into the intricacies of ICL and few-shot learning within the context of LLMs.

\subsubsection{ICL in LLMs}

ICL has gained significant attention with the rise of LLMs. 
ICL allows models to adapt to tasks based on contexts, without explicit gradient-based fine-tuning. 
A notable exemplar of this approach is GPT-3, which has demonstrated the capability to process tasks based on provided context \cite{brown2020language}. 
This paradigm shift challenges traditional methods that rely on task-specific fine-tuning.

Further insights into ICL suggest that during their pre-training phase, LLMs internalize extensive array of ``concepts'', positioning them within the domain of implicit Bayesian reasoning proposed by \citet{xie2021explanation}. 
This perspective is further enriched by studies that draw parallels between LLMs and topic models, proposing that LLMs infer latent conceptual variables during their inference phase \cite{wang2023large}. 

A more granular understanding of ICL is provided by \citet{pan2023incontext} with characterizing two ways through task recognition (TR) and task learning (TL). While TR pertains to discerning tasks from demonstrations, TL is more oriented towards assimilating new input-label associations. This distinction underscores the intricate relationship between TR and TL, emphasizing the enhanced efficacy of TL when combined with model scaling and enriched demonstrations.

\subsubsection{Few-shot Learning in LLMs}

Few-shot learning aims to enable models to learn new tasks using limited annotated examples. 
LLMs, especially those pre-trained on a large scale like BERT \cite{devlin2019bert}, have showcased a remarkable ability to efficiently learn new tasks through fine-tuning \cite{zhang2021revisiting}. 
A notable approach in this domain is pattern-exploiting training, which reformulates text classification tasks into cloze questions or ``prompts'' that resemble masked language modeling \cite{schick2020exploiting, schick2020its}. 
This approach has been further refined by techniques that automatically generate prompts and incorporate demonstrations into the input \cite{gao2020making}, and by methods that densify the supervision signal with label conditioning \cite{tam2021improving}. 
However, because the downstream tasks are often learned in isolation, it is still difficult to generalize to completely unseen tasks. 
\citet{ye2021crossfit} improves few-shot learning capabilities on unseen tasks by harnessing cross-task generalization ability from diverse known tasks.

\subsection{Dynamic Adapter Composition Techniques}

The ascent of Large Language Models (LLMs) has brought the necessity for agile and efficient adaptation techniques to the forefront, especially in light of the computational and memory demands inherent to these models. 
Two prominent strategies, Adapters and Mixture of Experts (MoE), have emerged as leading solutions in this space. 
While both champion modularity and adaptability, they offer distinct methodologies and advantages, which we explore in this section.

\subsubsection{Adapters}

Introduced by \citet{houlsby2019parameter}, adapters have revolutionized neural architectures. 
These modular units merge well with pre-trained models, facilitating adaptation to novel tasks without tampering with the original parameters. 
This paradigm shifts from conventional fine-tuning, which extensively modifies the entire model, to adapters, which streamline the adaptation process by curtailing computational overhead. Furthering the adapter concept, low-rank adaptation (LoRA) \cite{hu2021lora} freezes the base model parameters and trains a supplementary lightweight module instead.

Recent advancements in dynamic adapter composition include module decomposition and recomposition based on functionality \cite{kingetsu2021neural}, behavior modulation using task vectors \cite{ilharco2022editing}, and crafting modules through specific arithmetic operations \cite{zhang2023composing}. LoraHub \cite{huang2023lorahub} stands out by leveraging the modularity of LoRA modules, facilitating the assembly of diverse modules.




\subsection{Task Definition Extraction}

For LLMs, task definition extraction is paramount, ensuring they generalize effectively across a wide array of tasks. a primary mechanism for task definition is through instructions. 
By providing models with lucid and precise instructions, they can better interpret and execute diverse tasks. 
This approach emerges as a versatile alternative to the traditional supervised learning paradigm. 
The efficacy of finetuning LLMs using instruction-output pairs has been demonstrated, marking significant progress in this domain \cite{mishra2022cross, yin2022contintin, sanh2022multitask}. 

While the importance of high-quality instructions is undeniable, sourcing them often hinges on human expertise. This presents challenges in terms of scalability and uniformity. Some instruction sets, like OpenAssistant \cite{kopf2023openassistant}, are publicly available, but many remain proprietary. Addressing the intricacies of instruction generation, several innovative methodologies have been proposed. For instance, GPT-3 has been leveraged to derive instructions from initial seeds \cite{honovich2022unnatural} autonomously. In a similar vein, the Self-instruct method \cite{wang2022self} enables LLMs to generate both instructions and their associated outputs autonomously. 
Further enriching this domain, a novel technique has been introduced, which iteratively refines instruction prompts by integrating a seed model with a web corpus \cite{li2023self}.

Our methodology, TEGEE, aligns with the prevailing emphasis on instructions for task definition in LLMs. Unlike traditional methods focusing on autonomous instruction generation or refinement, TEGEE integrates task definition extraction with expert ensembling. This approach ensures optimal model performance by accurately interpreting task definitions and distinguishing TEGEE as a more efficient and adaptive solution in the LLM landscape.

\section{Methods}
TEGEE represents a model ensembling method for generalizable and continual few-shot learning.  This section delves into the details about how TEGEE performs few-shot learning. 

The first stage, \textbf{Expert Pool Construction}, focuses on generating task-based experts. These experts serve as knowledge candidates for our model. In the second stage, \textbf{Task Definition Extractor}, we introduce how to train a task definition extractor to conclude task definitions from given demonstrations. In the third stage, \textbf{Task Definition Guided Retriever}, the process of retrieving task experts based on input criteria is detailed. The fourth stage, \textbf{Expert Ensembling}, concentrates on ensembling tasks that correlate with the provided inputs. Finally, in the fifth stage, labeled \textbf{Continual Few-shot Learning} we explore the adaptability of our paradigm to support continual and lifelong learning, ensuring its flexibility and generalizability.

\subsection{Expert Pool Construction}
\label{sec:expert pool}
Within this section, our objective is to create an expert pool as our prior knowledge. Each expert specializes in a specific task and is trained with LoRA \cite{hu2021lora}. 
LoRA highlights the capability of pre-trained language models to exhibit a reduced dimension while maintaining their efficacy in training.  
Given \( W_0 \) from \( \mathbb{R}^{d \times k} \), modification occurs by expressing the subsequent update as a low-rank decomposition, formulated as \( W_0 + \Delta W = W_0 + BA \). 
Here, \( B \) belongs to \( \mathbb{R}^{d \times r} \), \( A \) is from \( \mathbb{R}^{r \times k} \), with the rank \( r \) being substantially lesser than the lesser of \( d \) and \( k \). 
Throughout the training phase, \( W_0 \) does not update by gradient. However, \( A \) and \( B \) have parameters open for adjustments.  Given \( h = W_0x \), the adapted forward pass is:
\begin{equation}
h = W_0x + \Delta W x = W_0x + BAx 
\end{equation}

 For each task within SuperNI, we train a distinct expert LoRA adaptor. These adaptors can be integrated into our existing models. Consequently, for each task, we not only train a dedicated expert but also establish a tuple encompassing the instruction for each task and its corresponding adaptor. Given the context, we denote our expert pool by \(\mathbf{E}\). Within \(\mathbf{E}\), we encompass \(N\) task units, each corresponding to a specific task from SuperNI. Each unit consists of a task definition, denoted as \(\textbf{T}_i\), and its associated trained adaptor, represented as \(\textbf{A}_i\).

\subsection{Task Definition Extractor}
\label{ref:extractor}
In this section, we train a specialized task definition extractor. Given the in-context learning (ICL) format $\{\textbf{I}_0, \textbf{L}_0, \ldots, \textbf{I}_k, \textbf{L}_k, \textbf{I}_{\text{final}}\}$, the goal for language models (LLMs) is to produce the final label $\textbf{L}_{\text{final}}$. Here, $\textbf{I}_i$ and $\textbf{L}_i$ represent the given demonstration pairs input and label for $i = 0, \ldots, k$, and $\textbf{I}_{\text{final}}$ is the final question for which we want the LLMs to respond with $\textbf{L}_{\text{final}}$. The parameter $k$ denotes the number of demonstrations. Our objective is to identify the format and task attribute similarities in the given demonstrations, ranging from $\textbf{I}_0, \textbf{L}_0$ to $\textbf{I}_k, \textbf{L}_k$, and to extract appropriate task definitions $\textbf{T}_{\text{concluded}}$ for them. Directly applying task definitions and demonstrations in ICL format into model training may result in issues with generalizing to out-of-distribution tasks. So, we align with the Stanford Alpaca framework \cite{alpaca}, add extra data in accordance with the Alpaca-GPT-4 instruction following data in \citet{peng2023instruction}. This approach enables our extractor to summarize suitable task definitions for provided demonstrations with proper instructions.

\subsection{Task Definition Guided Retriever}
\label{sec:retriever}

In this section, we elaborate on how to retrieve the relevant expert from the expert pool based on a given instruction. After acquiring an instruction, we input it into SentenceBERT\citep{reimers2019sentencebert}, resulting in an embedding \( E_{\text{ref}} \). Similarly, all the task definitions within the pool are also processed through Sentence-BERT to generate their respective embeddings \( E_i \). We then utilize the cosine similarity to compute the distances between \( E_{\text{ref}} \) and each \( E_i \). The formula for this distance metric is as follows: $ \text{Distance}(E_{ref},E_i) = \cos(E_{ref}, E_i) $ We subsequently select the top-\( k \) most similar adaptors based on these computed distances and normalize each distance from $w_0$ to $w_k$ as our following model ensemble weights.

\subsection{Expert Ensembling} 


Model Soup \citep{wortsman2022model} is an ensemble technique that enhances performance by merging multiple models to serve as the initialization for the final model's tuning. Inspired by this work, we consider our neural network defined as $\theta = \text{FineTune}(\theta_0, w)$, where $\theta_0$ represents the parameters to be initialized with the adaptors from our expert pool $\mathbf{E}$, and $w$ denotes the ensemble weights discussed in Section \ref{sec:retriever}. We employ a weighted average to initialize $\theta_0$ using the top-$k$ adaptors $\mathbf{A}_1, \ldots, \mathbf{A}k$ as follows:
\begin{equation}
\theta_0 = \frac{1}{W} \sum_{i \in S} w_i \textbf{A}_i, \quad \text{where} \quad W = \sum_{i \in S} w_i
\end{equation}
with $S \subseteq {1, \dots, k}$ being the set of indices corresponding to the top-$k$ adaptors adaptors as selected and discussed in Section \ref{sec:retriever}.

\subsection{Continual Few-shot Learning} 

After expert ensembling, we finetune our model using LoRA \cite{hu2021lora} with the weight initialization \(\theta_0\) and the given demonstrations $\{\textbf{I}_0, \textbf{L}_0, \ldots, \textbf{I}_k, \textbf{L}_k, \textbf{I}_{\text{final}}\}$. The final model parameters, \(\theta_{\text{final}}\), are obtained as follows:
\begin{equation}
\theta_{\text{final}} = \text{FineTune}(\theta_0, w_i),
\end{equation}
where \(\theta_{\text{final}}\) is used for inference on the final examples $\textbf{I}_{\text{final}}$ The benefits of our paradigm include the ability to handle demonstrations from any number of shots. Given a sufficient number of examples, we can further finetune the experts with these examples without limited token numbers like in ICL. Moreover, we can extend our expert pool \(\mathbf{E}\), with the finetuned adaptor parameter \(\mathbf{\theta}\) and the concluded task \(\textbf{T}_{\text{concluded}}\) from the task definition extractor (as described in \ref{ref:extractor}). To ensure diversity and uniqueness in expert pool $\mathbf{E}$, new task definitions $T_{\text{concluded}}$ are compared for similarity with existing entries $T$ in $\mathbf{E}$. If $T_{\text{concluded}}$ is sufficiently distinct (below a certain similarity threshold) from all entries in $\mathbf{E}$, it is added to the pool, thus updating $\mathbf{E}$ to include both diverse and unique definitions.

\section{Experimental Setup}
\subsection{Evaluating the Performance of  TEGEE}

\begin{table*}[htb]
\centering
\resizebox{1\textwidth}{!}{
\begin{tabular}{@{}lcc|cc|cc|cc|cc|cc|cc@{}}
  \toprule
  Category & \multicolumn{2}{c|}{\makecell{All}} & \multicolumn{2}{|c|}{\makecell{Coreference \\ Resolution}} & \multicolumn{2}{|c|}{\makecell{ Data \\ to Text}} & \multicolumn{2}{|c|}{\makecell{ Answerability \\Classification}} & \multicolumn{2}{|c|}{\makecell{ Question \\ Rewriting}} & \multicolumn{2}{|c|}{\makecell{ Title \\ Generation}} & \multicolumn{2}{|c}{\makecell{ Word \\ Analogy}} \\\midrule
   & K = 5 & K = 500 & K = 5 & K = 500& K = 5 & K = 500& K = 5 & K = 500& K = 5 & K = 500& K = 5 & K = 500& K = 5 & K = 500 \\\midrule
        T5-small &24.95 &52.88 & 23.20& 53.49& 16.98& 34.33& 30.92& 67.56& 47.58& 60.69& 15.74& 30.34& 8.68& 58.89 \\\midrule
        T5-base &37.17&64.75& 32.13& 65.82& 26.05& 40.42& 55.10& 77.46& 55.13& 63.52& 22.82& 36.53& 12.04& 99.94 \\\midrule
        T5-large &40.07&69.84& 39.72& 76.57& 30.66& 43.45& 62.02& 81.10& 50.15& 65.07& 26.56& 38.85& 17.98& 100.00\\\midrule
        T5-3B&46.81&73.52&51.78& 83.61& 35.93& 46.07& 63.79& 80.77& 59.70& 66.34& 30.42& 42.06& 24.10& 100.00 \\\midrule
\toprule       
Category & \multicolumn{2}{c|}{\makecell{All}} & \multicolumn{2}{|c|}{\makecell{ Textual \\ Entailment}} & \multicolumn{2}{|c|}{\makecell{ Cause Effect \\ Classification}} & \multicolumn{2}{|c|}{\makecell{ Grammar \\ Error Correction}} & \multicolumn{2}{|c|}{\makecell{ Overlap \\ Extraction}} & \multicolumn{2}{|c|}{\makecell{ Keyword \\ Tagging}} & \multicolumn{2}{|c}{\makecell{ Dialogue \\ Act  Recognition}} \\\midrule
 & K = 5 & K = 500 & K = 5 & K = 500& K = 5 & K = 500& K = 5 & K = 500& K = 5 & K = 500& K = 5 & K = 500& K = 5 & K = 500 \\\midrule
        T5-small &24.95 &52.88& 22.95& 59.10& 35.63& 53.84& 62.94& 78.4& 17.62& 59.07& 25.55& 50.39& 26.18& 62.83  \\\midrule
        T5-base &37.17&64.75& 43.79& 71.71& 43.45& 65.55& 73.19& 83.55& 41.76& 67.86& 32.08& 57.49& 35.69& 83.51 \\\midrule
        T5-large &40.07&69.84& 42.52& 80.17& 49.08& 74.02& 69.42& 83.39& 36.44& 70.07& 36.03& 61.19& 39.33& 89.62 \\\midrule
        T5-3B &46.81&73.52 &50.28& 88.04& 53.63& 79.04& 82.99& 85.96& 43.72& 73.30& 43.39& 64.70& 47.58& 90.45 \\\bottomrule
\end{tabular}
}
\caption{The performance of TEGEE across different task categories. K represents the number of samples. Due to space constraints, we select scores from both the 5-sample and the 500-sample settings for result demonstration. For more details, please refer to Appendix Table  \ref{tab:categories all results}.}
\label{tab:categories ablation}
\end{table*}
We conduct experiments based on SuperNI, a dataset comprised of human-authored instructions, which encompasses a variety of tasks, to verify the generalization and performance enhancement of TEGEE.
We point out that only the parameters of the task definition extractor, task definition guided retriever, and ensembled expert are activated during the inference pipeline of TEGEE.
The estimated activated parameter size of TEGEE is around 6B if we utilize our 3B paradigm.
As a result, we select LLaMA2-7B, LLaMA2-13B, and Falcon-7B as our baselines, which the most widely utilized open-source models, demonstrating that TEGEE achieves comparable few-shot performance to LLMs twice its size.

In this section, we detail the SuperNI-based experiment setting. Firstly, we construct an expert pool. For this purpose, we employ the T5 model series \citep{2020t5} as our expert base model. 
Each T5 expert is specifically trained with LoRA \citep{hu2021lora} for each task from SuperNI official training splits to constitute TEGEE's prior knowledge. 
Due to many tasks being smaller in size, we opt for a 9:1 split for training and validation. 
Had we gone for an 8:2 train-test split, our overall training set would have been insufficient. 
For each task, adaptors and instructions are extracted from SuperNI. 
During training, the batch size is set to 40, the learning rate at 3e-4, the input length is 1024, and the LoRA trained parameters are the "q\_proj" and k\_proj modules in the multi-head attention modules.

Subsequently, we train a Task Definition Extractor using the Galactica 3B model. 
Within this framework, we utilized Alpaca-GPT-4 \cite{peng2023instruction} and 20K samples from SuperNI \cite{supernaturalinstructions} for training. From each task, 200 samples were chosen, amounting to 15K in total. 
This shortfall is due to some tasks having less than 200 samples. To meet the requirement of 20K samples, we sampled more from tasks with a larger sample size. 
The model's training follows the guidelines set by Stanford Alpaca, with a batch size of 32 and full parameter training.

During the Continual Training phase, we conduct a final round of continual learning, concentrating on the specified Input and Output. 
For model training, we utilized LoRA (as described by \citet{hu2021lora}), focusing on the "q\_proj" and "k\_proj" modules within the multi-head attention mechanisms. The rank for LoRA was set to 4, the learning rate was established at 3e-4, and the input length was fixed at 1024.

\subsection{Ablation Experiment}

We conduct two ablation experiments to investigate the roles of the task definition and expert ensemble components in the TEGEE process. 
We reconstructed the input-output data of demonstrations and created two formats of labeled data to train LoRA in two different ways.

The setting of TEGEE without definitions and ensemble, as stated in Table~\ref{tab:ablation}, refers to directly using few-shot input-output pairs as labeled data to train randomly initialized experts.
In this scenario, the TEGEE stage lacks both task definitions derived from user demonstrations and expert ensembling and degenerates into few-shot learning. 
We conduct these ablation experiments on settings with 5 samples, 50 samples, 100 samples, and 500 samples, as well as using T5 small, T5 base, T5 large, and T5 3B models, to investigate the impact of task definition in TEGEE under different data scales and model sizes.

The ablation experiment setting of TEGEE without ensemble firstly retrieves task definition from user demonstrations using our own 3B extractor. 
Next, we add the task definition into the input field in the user demonstrations and construct input-output pairs as labeled data for training LoRA. 
Compared to standard TEGEE, TEGEE without ensembling lacks the initialization of LoRA weights through expert ensembling and trains LoRA from scratch.
We conduct the same experiments as TEGEE without definitions and ensembling to explore the influence of expert ensembling in TEGEE under different data scales and model sizes.

\subsection{Analysing Task Definition Quality}
We propose two methods to analyze the quality of task definitions concluded from user demonstration. 
One is an automatic quality analysis approach, and the other is a human analysis approach.

TEGEE retrieves from the expert pool using the generated task definition and matches the top 3 experts based on similarity. 
This similarity is computed by using the task definition generated from user input and the task definitions stored in the expert pool. Automatic quality analysis approach counts the number of overlapping matched experts as a measure of the commonality of task definitions from different sources, such as task definitions generated by GPT-3, GPT-3.5,  GPT-4, our extractor, and those stored in the expert pool.

We conduct human analysis on the quality of generated task definitions, comparing human task definitions with those generated by the model, and categorize them into the following four types: same, relevant task with missing details, relevant task with misleading information, different.
Each sample in the human analysis is annotated by one of the authors, instead of crowd-sourcing annotators, to guarantee the validity of the human analysis.

\section{Results}

\begin{figure}[tb]
    \centering
    \includegraphics[width=1\linewidth]{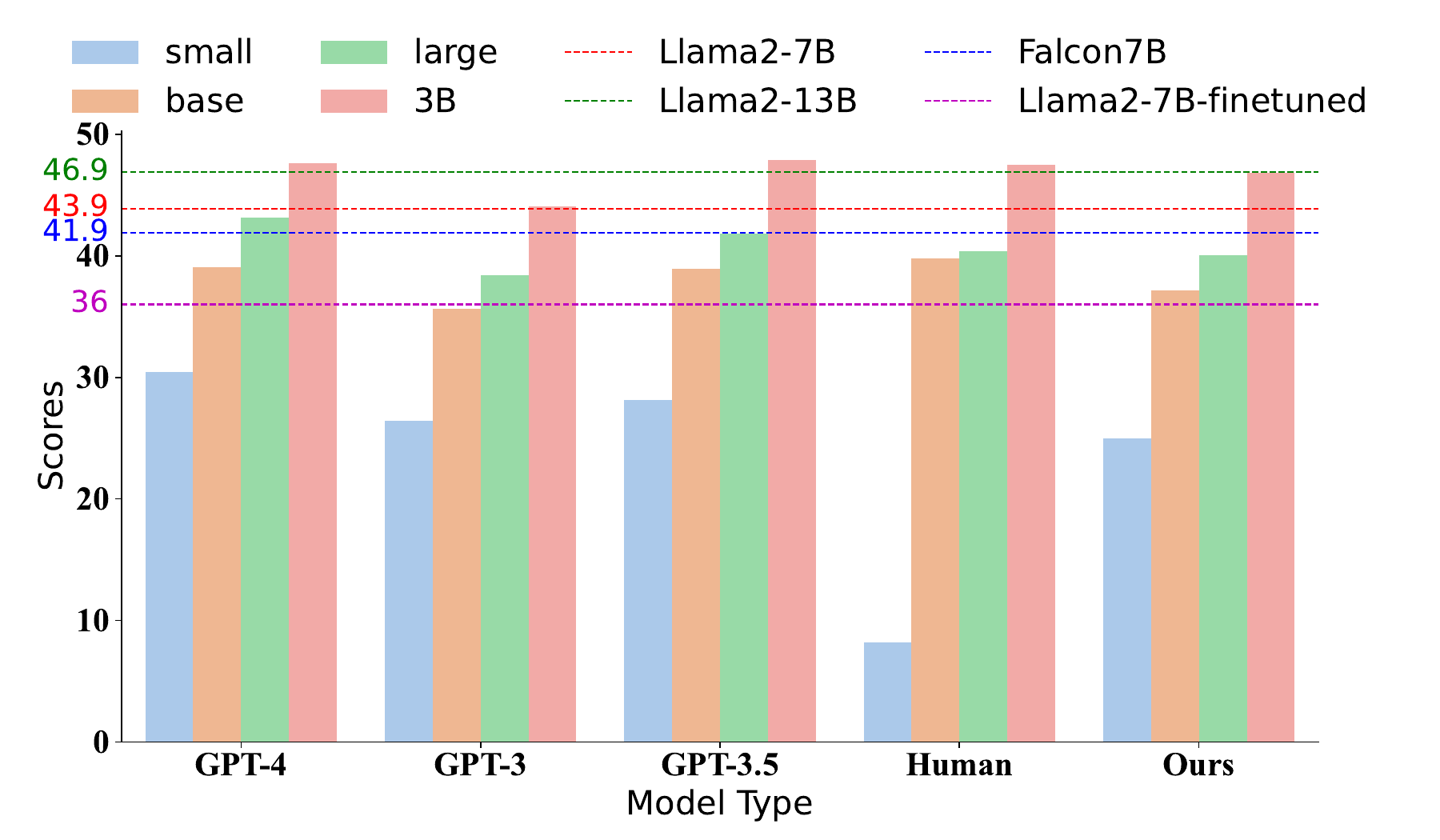} 
    \caption{Performance comparison of TEGEE across four distinct expert base model sizes and five task definition extractors with 5 demonstration inputs. The expert base models include LLaMA2-7B and LLaMA2-13B, and Falcon7B. Baseline ICL results are denoted by dashed lines. Notably, our paradigm employing a 6B parameters demonstrates superior performance to the LLaMA2-7B ICL and exhibits comparable results to the LLaMA2-13B ICL.} 
    \label{fig: deepicl 5shots}
\end{figure}

\subsection{Performance of TEGEE}

As shown in Figure \ref{fig: deepicl 5shots}, our work introduces a novel task definition extractor, outperforms the LLaMA2-7B ICL while slightly trailing behind the LLaMA2-13B ICL outcomes. Futhermore, GPT-4 surpasses human-annotated results, particularly in contexts involving smaller model sizes. 
This result can be explained by the preference of human annotations to incorporate excessive detail into task definitions. Compared to GPT-4 short instructions, too many excessive constraints would mislead the attention of models. 
Despite our task definition extractor operating on a 3B parameter framework, it outperforms the GPT-3 model, which is based on a 175B parameter setup. 
This success is amplified by the application of supervised fine-tuning strategies, allowing our task generator extraction methodology to achieve enhanced results compared to those obtained through traditional GPT-3 implementations.

\begin{figure*}[htb]
    \centering
    \includegraphics[width=\textwidth]{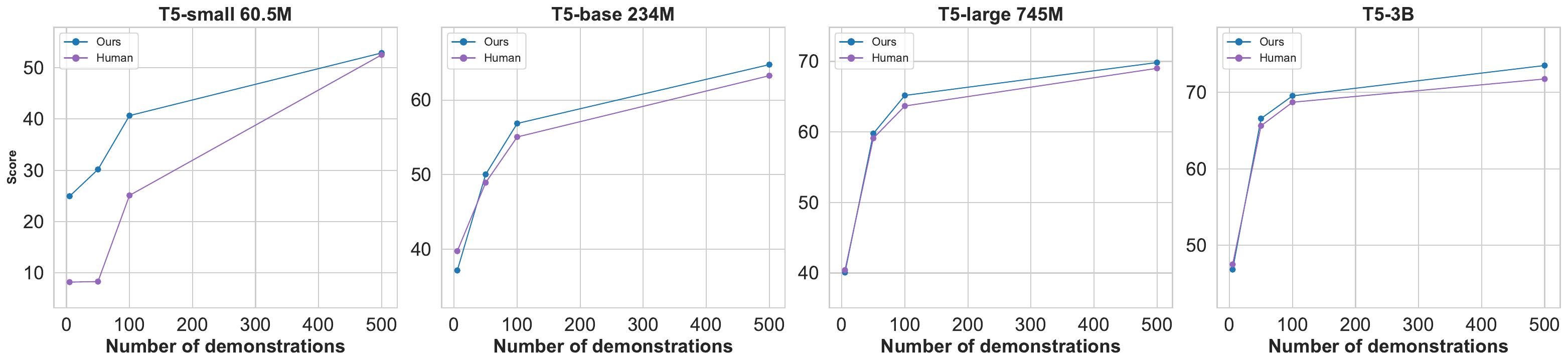}
    \caption{The Performances trend alongside increases in the number of demonstrations.}
    \label{fig:trends}
\end{figure*}

According to Figure \ref{fig:trends} and \ref{tab:categories all results}, the increasing number of input demonstrations can significantly improve the accuracy of our models in each following categories, especially in smaller-sized configurations. 
For larger models, we observe a substantial performance improvement when increasing from 5 to 100 demonstrations. 
However, beyond 100 demonstrations, the rate of improvement in our model's performance begins to plateau, indicating a more steady increase. 

\subsection{Ablation Study}

\begin{table}[htb]
\centering
\resizebox{0.47\textwidth}{!}{
\begin{tabular}{@{}lllll@{}}
  \toprule
  Size & 5 samples & 50 samples & 100 samples  & 500 samples \\\midrule
  \multicolumn{5}{c}{\textbf{without definitions and ensembling} ({\color{olive}-44.71})} \\
        small & 7.91 ({\color{olive}-17.04}) & 7.79 ({\color{olive}-22.38}) & 6.93 ({\color{olive}-33.73}) & 7.41 ({\color{olive}-45.47}) \\
        base & 9.91 ({\color{olive}-27.26}) & 9.79 ({\color{olive}-40.25}) & 8.98 ({\color{olive}-47.88}) & 8.51 ({\color{olive}-56.24}) \\
        large & 6.92 ({\color{olive}-33.15}) & 7.75 ({\color{olive}-52.03}) & 8.16 ({\color{olive}-57.02}) & 8.63 ({\color{olive}-61.21}) \\
        3B & 8.57 ({\color{olive}-38.24}) & 8.67 ({\color{olive}-57.90}) & 8.91 ({\color{olive}-60.64}) & 8.57 ({\color{olive}-64.95}) \\\midrule
  \multicolumn{5}{c}{\textbf{without ensembling} ({\color{cyan}-4.61})} \\
        small & 13.51 ({\color{cyan} -11.44}) & 17.10 ({\color{cyan} -13.07}) & 37.10 ({\color{cyan} -3.56}) & 51.50 ({\color{cyan} -1.38}) \\
        base & 26.40 ({\color{cyan} -10.77}) & 47.10 ({\color{cyan} -2.94}) & 54.40 ({\color{cyan} -2.46}) & 64.59 ({\color{cyan} -0.16}) \\
        large & 25.32 ({\color{cyan} -14.75}) & 57.12 ({\color{cyan} -2.66}) & 63.27 ({\color{cyan} -1.91}) & 68.88 ({\color{cyan} -0.96}) \\
        3B & 41.16 ({\color{cyan} -5.65}) & 65.58 ({\color{cyan} -0.99}) & 68.76 ({\color{cyan} -0.79}) & 73.33( {\color{cyan} -0.19}) \\\midrule\midrule
\multicolumn{5}{c}{\textbf{TEGEE} (w/ Trained \textbf{TDG})} \\
        small & \multicolumn{1}{c}{24.95} & \multicolumn{1}{c}{30.17} & \multicolumn{1}{c}{40.66} & \multicolumn{1}{c}{52.88} \\
        base & \multicolumn{1}{c}{37.17} & \multicolumn{1}{c}{50.04} & \multicolumn{1}{c}{56.86} & \multicolumn{1}{c}{64.75} \\
        large & \multicolumn{1}{c}{40.07} & \multicolumn{1}{c}{59.78} & \multicolumn{1}{c}{65.18} & \multicolumn{1}{c}{69.84} \\
        3B & \multicolumn{1}{c}{46.81} & \multicolumn{1}{c}{66.57} & \multicolumn{1}{c}{69.55} & \multicolumn{1}{c}{73.52} \\\bottomrule
\end{tabular}
}
\caption{Ablation results. We conduct ablation experiments on T5\_small, T5\_base, T5\_large, and T5\_3B models, using 5 samples, 50 samples, 100 samples, and 500 samples. {\color{olive}Olive color values} and {\color{cyan} cyan color values} indicate ablations of TEGEE without definitions and ensembling and TEGEE without ensembling, respectively.
\textbf{TDG} refers to \textit{\textbf{task definition generator}}.}
\label{tab:ablation}
\end{table}

Compared to the final TEGEE, TEGEE without definitions and ensembling exhibits an average performance decrease of 44.71\% across various data scales and model sizes.
This underscores the significant role of task definitions in TEGEE, particularly within demonstrations of the same task type.
In addition, another observation is that TEGEE without definitions and ensembling does not show significant performance improvements as the data scale and model size increases, indicating that without explicit (ours) or implicit (ICL) task definition extraction, few-shot learning does not work. 
From another perspective, it underscores the crucial role of task definitions for this paradigm, both in ICL and in our approach.

In the ablation experiment setting of TEGEE without ensembling, there are four experimental setups ([T5 small, 5 samples], [T5 small, 50 samples], [T5 base, 5 samples], [T5 large, 5 samples]) that perform more than 10\% worse than the final TEGEE. 
This indicates that in scenarios with small models or limited data, experts ensemble in TEGEE facilitates rapid bootstrapping of inference for related tasks.

As the number of samples increases and the model size grows, the performance gap between final TEGEE and TEGEE without ensembling gradually diminishes. 
This suggests that the initialization of TEGEE by using expert ensembling does not negatively impact the effectiveness of model finetuning in the future.



\subsection{Task Definition Quality Analysis}

\subsubsection{Automatic Quality Analysis}
\label{sec:Automatic Quality Analysis}

\begin{table}[htb]
\centering
\renewcommand{\arraystretch}{1.2}
\resizebox{0.47\textwidth}{!}{
\begin{tabular}{@{}cccccc@{}}
    \toprule
    \makecell{\textbf{Generator}}  & \textbf{GPT3} & \textbf{GPT-3.5} & \textbf{GPT-4} & \textbf{Human} & \textbf{Ours} \\ \midrule
    \textbf{GPT-3} & 351 & 25 & 16 & \underline{14} & 10 \\
    \textbf{GPT-3.5} & 25 & 351 & \textbf{98} & \underline{55} & 41 \\
    \textbf{GPT-4} & 16 & 98 & 351 & \underline{76} & 40 \\
    \textbf{Human} & 14 & 55 & 76 & 351 & 30 \\ 
    \textbf{Ours} & 10 & 41 & 40 & \underline{30} & 351 \\\bottomrule
\end{tabular}}

\caption{The number of overlapping matched experts. The \textbf{bold value} represents the maximum count of overlapping matched experts between task definitions from different sources, while the \underline{underlined values} represent the number of overlapping matched experts between human task definitions and other task definitions.}
\label{tab:confusion matrix}
\end{table}


From Table \ref{tab:confusion matrix}, we observe that the overlap between the experts retrieved by the task definitions generated by our 3B generator and those retrieved by human task definitions is 30. This value is higher than the corresponding value of 14 in the GPT-3 experimental setting but lower than the corresponding values of 55 and 76 in the GPT-3.5 and GPT-4 experimental settings, respectively. 
This may be due to the lack of instruction fine-tuning in GPT-3, resulting in poorer instruction-following capabilities and thus generating lower-quality task definitions.
In addition, GPT-3.5 and GPT4 match the highest number of the similar experts, which may be due to the semantic similarity in the task definitions they generate. Figure \ref{appendix: statics of similar experts} illustrates the statistics of overlapping similar experts matched by pairwise task definition in detail.
\subsubsection{Human Analysis}
\label{sec: Human Analysis}



\begin{table}[!tbp]
\small
\centering
\resizebox{0.93\columnwidth}{!}{
\begin{tabular}{@{}lcccc@{}}
\toprule
    \makecell{\textbf{Number}} & \textbf{Same} & \textbf{Sim\_1} & \textbf{Sim\_2} & \textbf{Different} \\\midrule
    \textbf{GPT-3} & 0 & 10 & 1 & 106 \\
    \textbf{GPT-3.5} & 30 & 29 & 24 & 34 \\
    \textbf{GPT-4} & 47 & 51 & 15 & 4 \\\midrule
    \textbf{Ours} & 9 & 22 & 19 & 67 \\
\bottomrule
\end{tabular}%
}
\caption{The number of categories. sim\_1 corresponds to the category of relevant task missing details, while sim\_2 corresponds to the category of relevant task with misleading information.}
\label{tab:human analysis}
\end{table}



Table \ref{tab:human analysis} reveals that the number of definitions generated by our 3B model that are the same or similar to the human task definition exceeds those of GPT3. But there is still a distance between our trained task definition extractor and GPT-3.5 and GPT4. 
This finding also corresponds to the conclusions of the automated quality analysis.
GPT-3.5 and GPT4 based task definitions guided retrievers lead to lightweight but consistently better performance compared to our trained task generator.
Please refer to Fig \ref{fig:human_analysis}  for more details on manual analysis.
\section{Conclusion}
In this paper, we present TEGEE, a new approach that addresses the limitations of existing ICL frameworks related to fixed input lengths and limited learning capabilities. TEGEE utilizes an expert pool to provide a strong knowledge base and improve learning from input demonstrations. We divide the ICL process into task definition extraction and task processing. Our findings show that accurately extracting task definitions is vital for ICL success. During task processing, we employ a novel initialization technique based on task definitions, leveraging insights from the expert pool, resulting in significantly enhanced performance and advancing the ICL process.
\section*{Limitations}
While our paradigm can alleviate the challenges associated with lifelong learning for Large Language Models (LLMs), the expansion of our expert pool with the addition of varied task definitions is inevitable. Nonetheless, a potential overlap in expertise within this pool may arise due to similarities between some experts. Addressing the optimization of this expert pool to manage such redundancies will be an essential focus of future work. In this section, an explicit task definition is extracted. Yet, whether a more effective approach could be achieved with solely implicit embeddings remains a subject for further discussion.
\section*{Ethics Statement}
This paper does not raise any ethical concerns. The data and additional resources employed in this study are open-source and widely utilized in existing works.

\bibliography{inscom}

\appendix

\section{Auto Quality Analysis Appendix}
\label{sec:appendix}

\begin{figure*}[htbp]
	\centering
	\begin{minipage}{0.32\linewidth}
		\centering
		\includegraphics[width=0.9\linewidth]{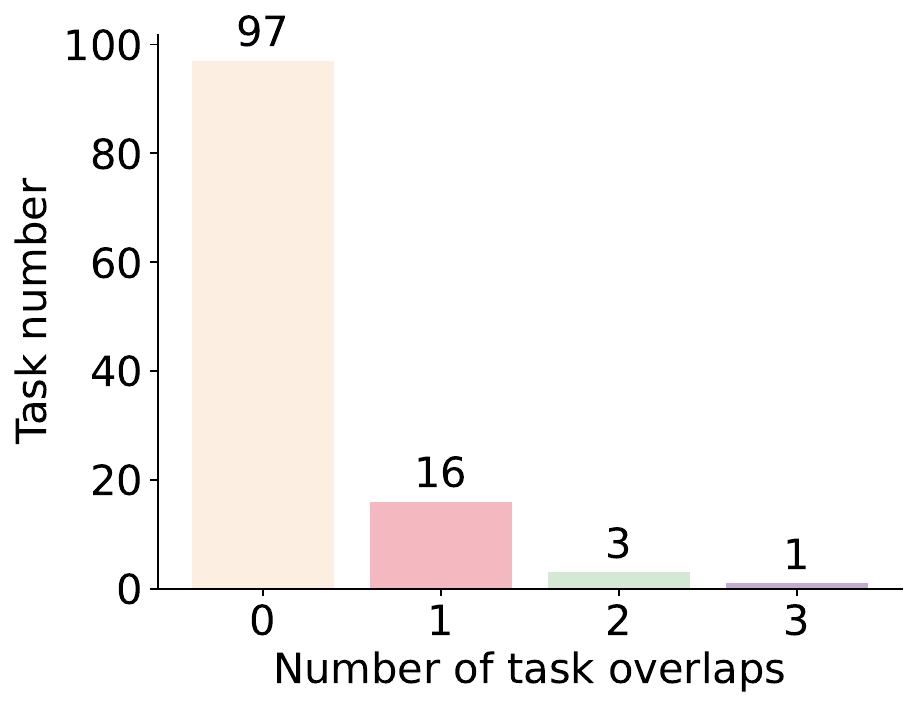}
		\subcaption{GPT3 \& GPT-3.5}
		\label{1}
	\end{minipage}
	\begin{minipage}{0.32\linewidth}
		\centering
		\includegraphics[width=0.9\linewidth]{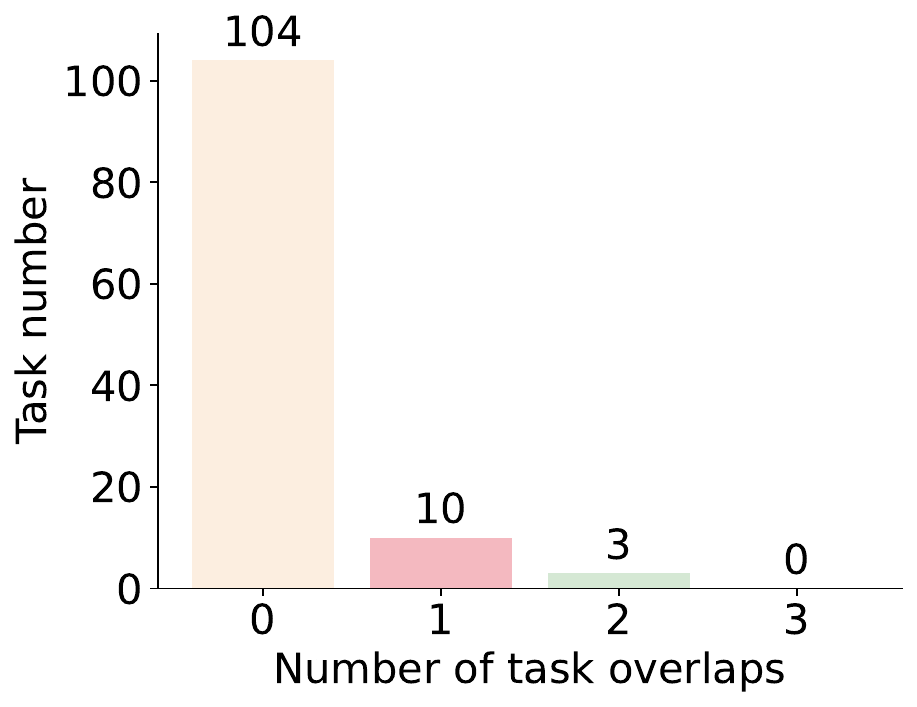}
		\subcaption{GPT3 \& GPT4}
		\label{2}
	\end{minipage}
  	\begin{minipage}{0.32\linewidth}
		\centering
		\includegraphics[width=0.9\linewidth]{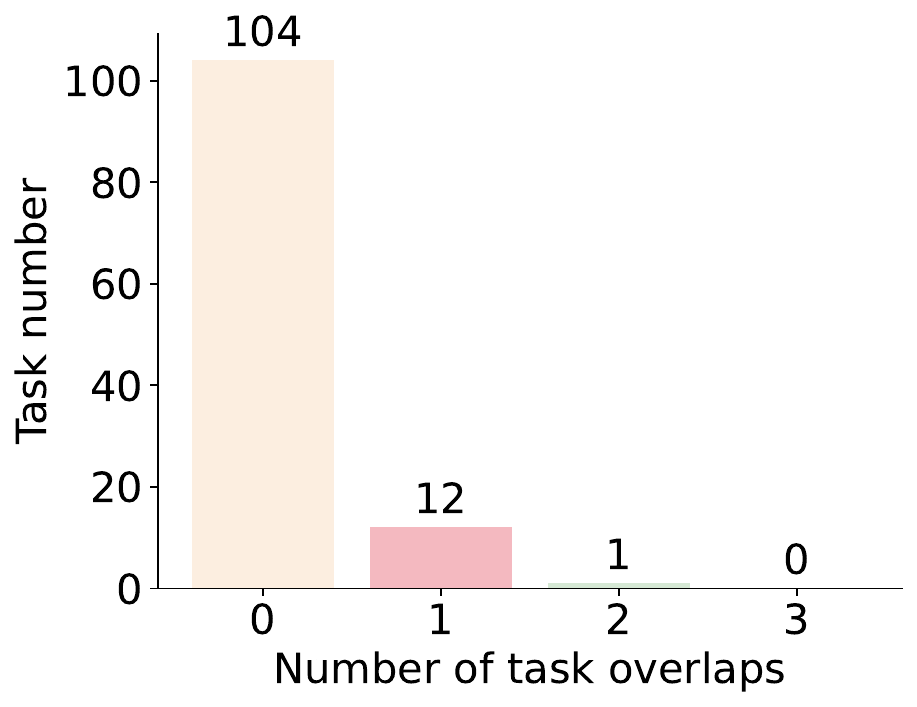}
		\subcaption{GPT3 \& Human}
		\label{3}
	\end{minipage}
        
        \medskip
 
        \begin{minipage}{0.32\linewidth}
		\centering
		\includegraphics[width=0.9\linewidth]{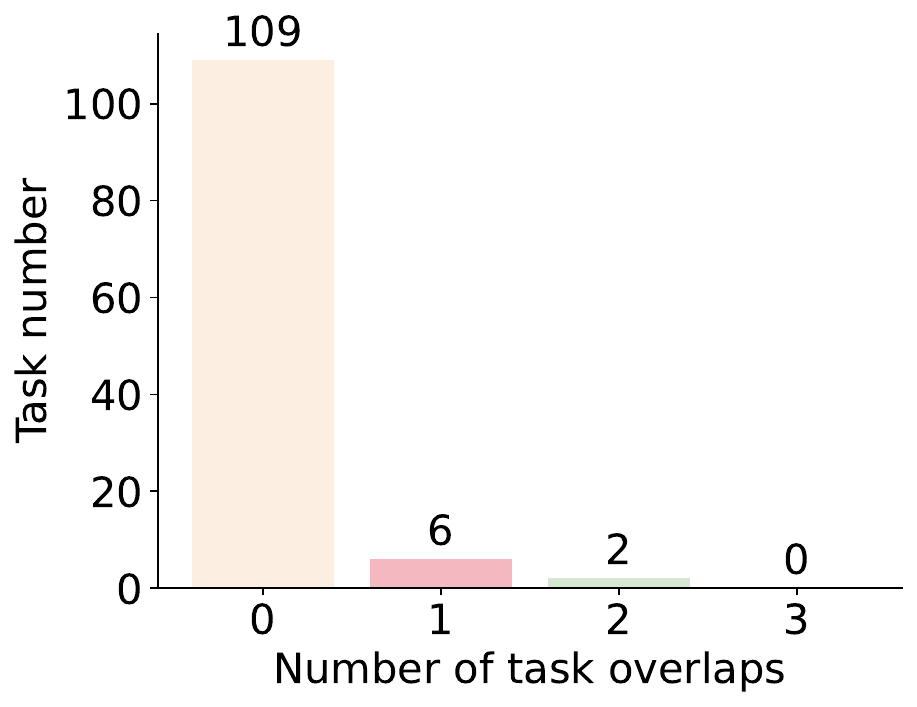}
		\subcaption{GPT3 \& Ours}
		\label{4}
	\end{minipage}
	\begin{minipage}{0.32\linewidth}
		\centering
		\includegraphics[width=0.9\linewidth]{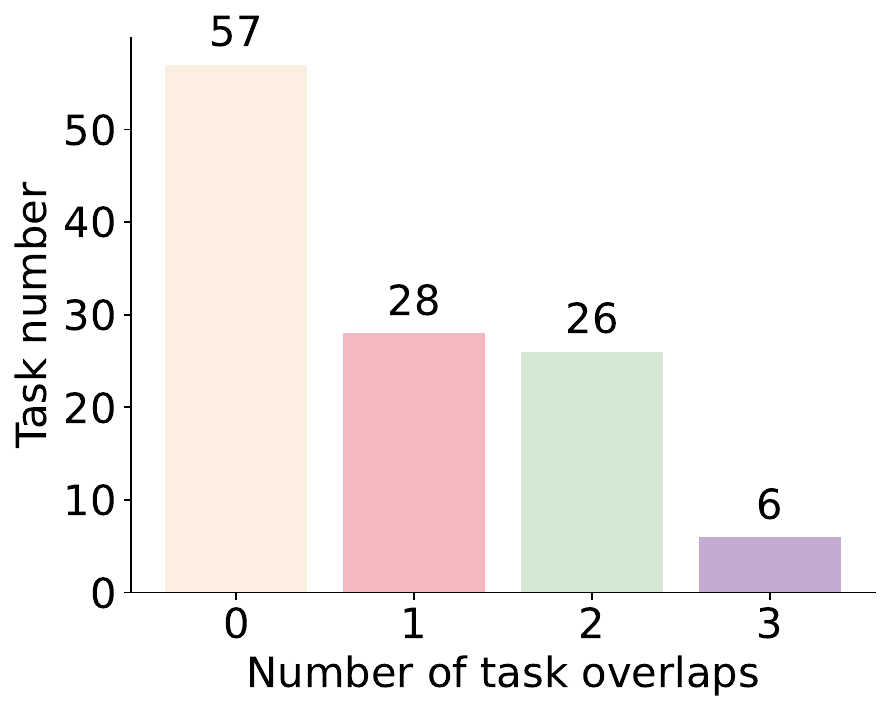}
		\subcaption{GPT-3.5 \& GPT4}
		\label{5}
	\end{minipage}
         \begin{minipage}{0.32\linewidth}
		\centering
		\includegraphics[width=0.9\linewidth]{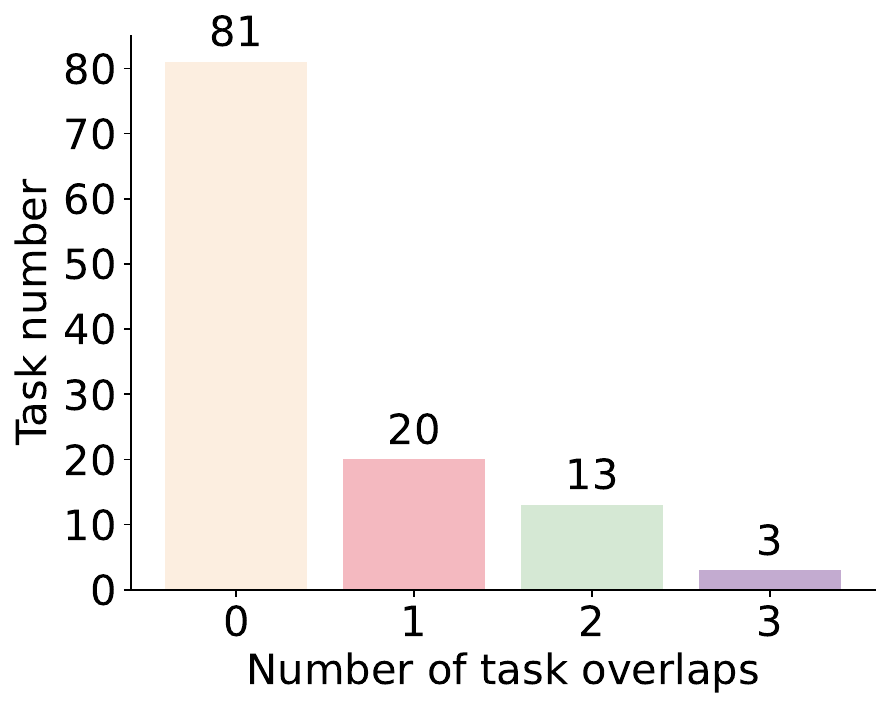}
		\subcaption{GPT-3.5 \& Human}
		\label{6}
	\end{minipage}
        \medskip

        \begin{minipage}{0.32\linewidth}
		\centering
		\includegraphics[width=0.9\linewidth]{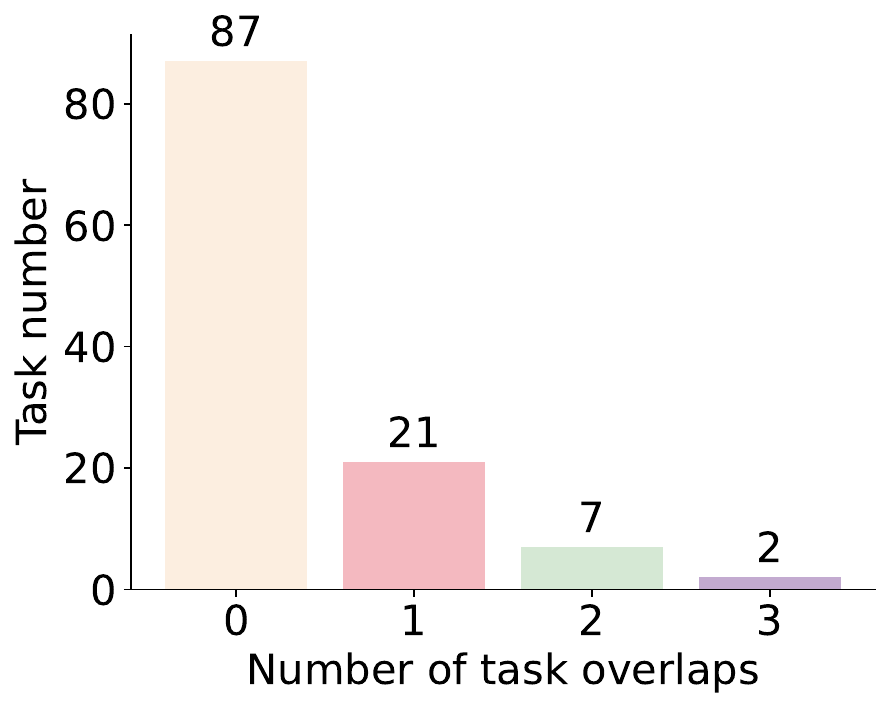}
		\subcaption{GPT-3.5 \& Ours}
		\label{7}
	\end{minipage}
	\begin{minipage}{0.32\linewidth}
		\centering
		\includegraphics[width=0.9\linewidth]{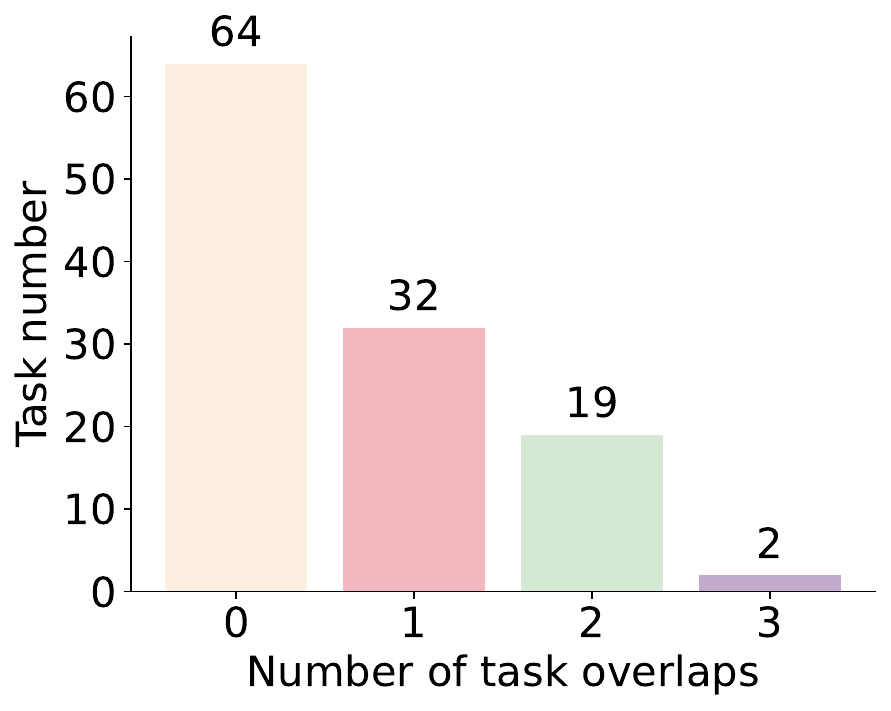}
		\subcaption{GPT4 \& Human}
		\label{8}
	\end{minipage}
         \begin{minipage}{0.32\linewidth}
		\centering
		\includegraphics[width=0.9\linewidth]{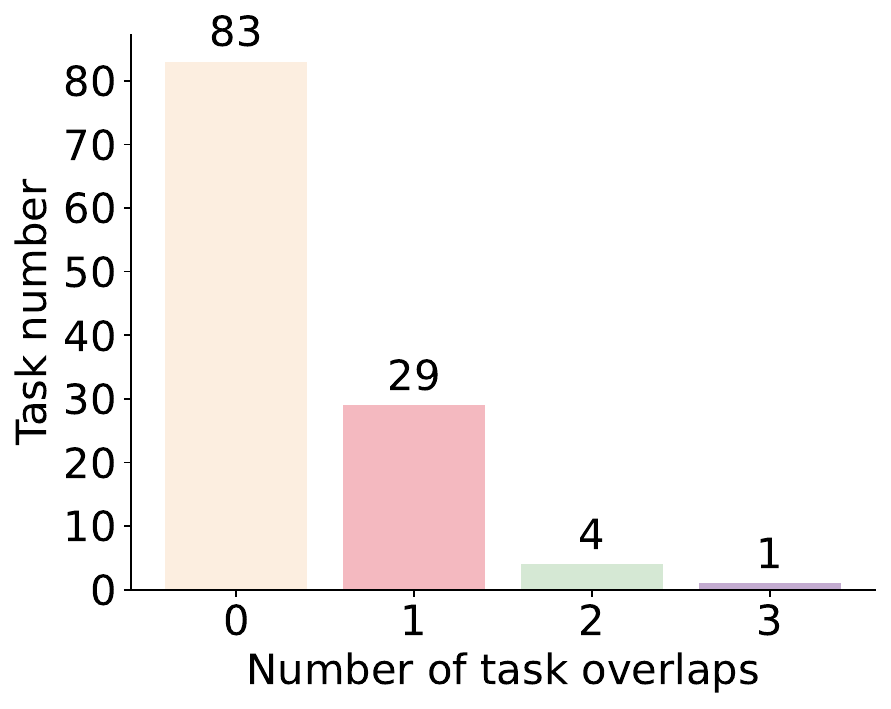}
		\subcaption{GPT4 \& Ours}
		\label{9}
	\end{minipage}
        \medskip

        \begin{minipage}{0.32\linewidth}
		\centering
		\includegraphics[width=0.9\linewidth]{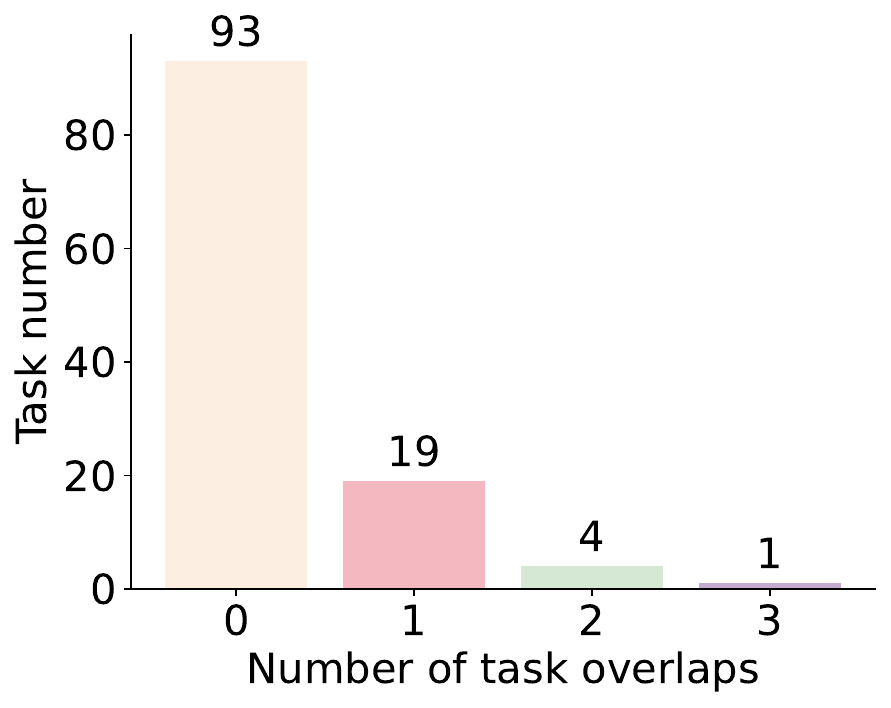}
		\subcaption{Ours \& Human}
		\label{10}
	\end{minipage}

\caption{The statistics of overlapping similar tasks matched by pairwise task definition. The horizontal axis represents the overlap count of similar tasks, while the vertical axis represents the number of matched tasks.}
\label{appendix: statics of similar experts}
\end{figure*}

Figure \ref{appendix: statics of similar experts} illustrates the statistics of overlapping similar tasks matched by pairwise task definition., as detailed in Sections \ref{sec:Automatic Quality Analysis}.
We recorded the overlap count of experts retrieved for pairwise task definition queries and depicted them in a bar chart.

\section{Examples in Human Analysis} 
\begin{figure*}[htb]
    \centering
    \includegraphics[width=1\textwidth]{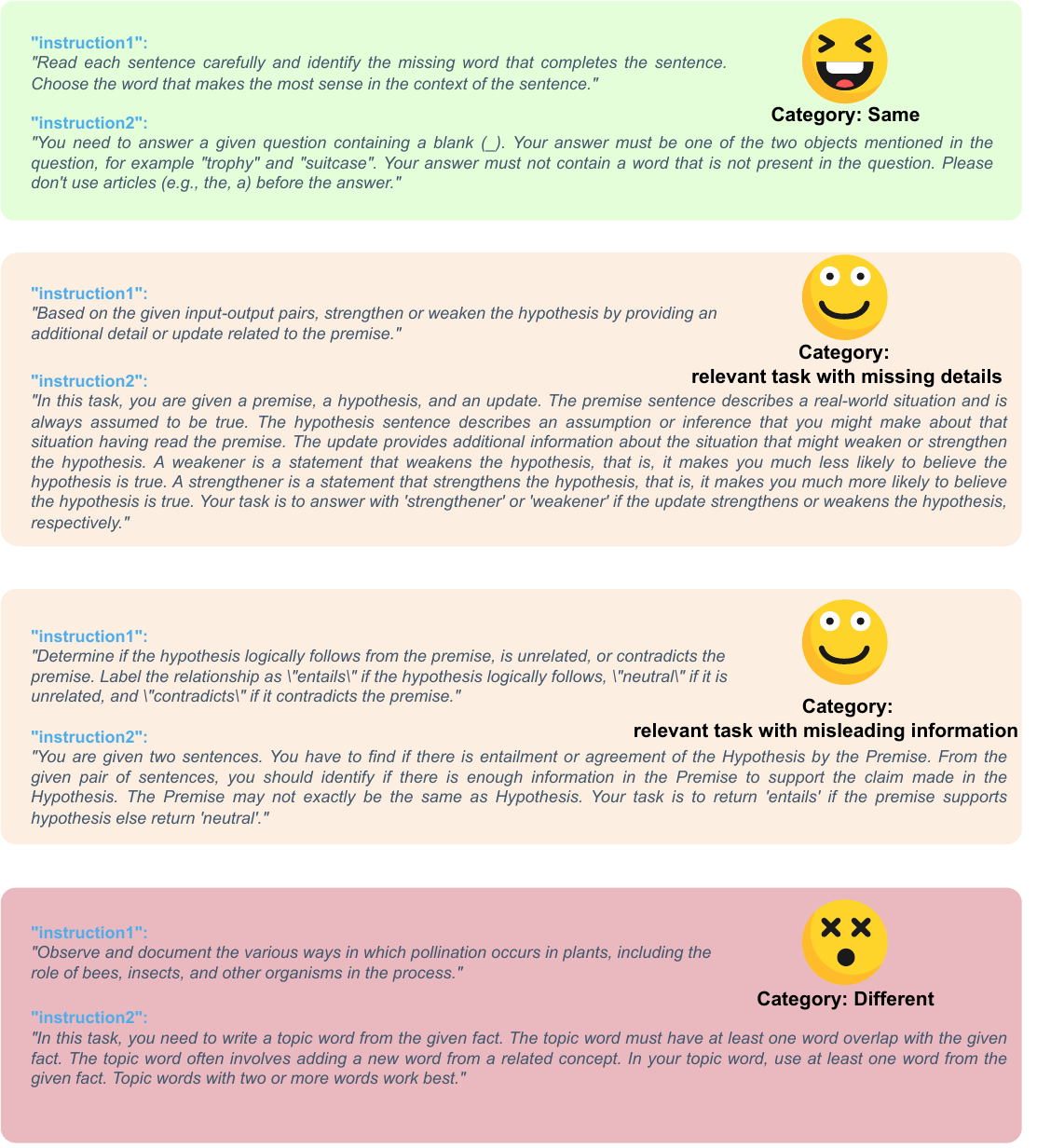}
    \caption{Examples of human analysis. We manually analyze the semantic similarity of paired instructions and categorize the samples into the following four classes: same, relevant task with missing details, relevant task with misleading information and different.}
    \label{fig:human_analysis}
\end{figure*}

Figure \ref{fig:human_analysis} illustrates specific examples from our human evaluation process, as detailed in Sections \ref{sec: Human Analysis}. We categorize the samples into the following four classes: same, relevant task with missing details, relevant task with misleading information and different.

\section{The Table of Model Performance} 
Table \ref{tab:accents} shows the performance of TEGEE under different experimental settings, such as task definition generator, LoRA
base model, and number of ICL examples. 
The human setting is to directly use the task definitions in the dataset. 

\section{Model Performance Across Different Categories.}
Table \ref{tab:categories all results} displays the performance of TEGEE across different task categories in detail.

\begin{table*}[htb]
\centering
\begin{tabular}{lccccc}
  \toprule
  Task Definition Generator & LORA base & Sample-5 & Sample-50 & Sample-100 & Sample-500 \\ \midrule
  \multirow{4}{*}{GPT3} & small & 25.70 & 31.46 & 41.91 & 52.36 \\
                        & base & 35.13 & 48.03 & 54.50 & 63.91 \\
                        & large & 38.82 & 56.99 & 62.97 & 68.99 \\
                        & 3B & 44.74 & 64.74 & 69.03 & 73.51 \\         \midrule
  \multirow{4}{*}{GPT-3.5} & small & \underline{29.06} & \underline{32.33} & \textbf{42.44} & \textbf{53.72} \\
                        & base & \underline{38.36} & \textbf{50.17} & \textbf{57.20} & 64.49 \\
                        & large & \underline{41.77} & \textbf{60.32} & \underline{65.81} & \underline{70.61} \\
                        & 3B & \textbf{47.95} & \textbf{66.78} & \textbf{70.22} & \textbf{74.24} \\       \midrule
  \multirow{4}{*}{GPT4} & small & \textbf{30.47} & \textbf{36.11} & \underline{42.22} & \underline{52.14} \\
                        & base & \textbf{39.09} & 49.59 & 56.67 & \underline{64.74} \\
                        & large & \textbf{43.14} & \underline{60.27} & \textbf{66.00} & \textbf{70.68} \\
                        & 3B & \underline{47.59} & 66.53 & \underline{69.84} & \underline{73.79} \\       \midrule
  \multirow{4}{*}{Ours} & small & 24.95 & 30.17 & 40.66 & 52.88 \\
                        & base & 37.17 & \underline{50.04} & \underline{56.86} & \textbf{64.75} \\
                        & large & 40.07 & 59.78 & 65.18 & 69.84 \\
                        & 3B & 46.81 & \underline{66.57} & 69.55 & 73.52 \\      \midrule \midrule
  \multirow{4}{*}{Human} & small & 8.20 & 8.31 & 25.10 & 52.51 \\
                        & base & 39.76 & 48.93 & 55.05 & 63.26 \\
                        & large & 40.41 & 59.11 & 63.68 & 69.02 \\
                        & 3B & 47.48 & 65.62 & 68.71 & 71.75 \\      \midrule  \midrule
  \multirow{4}{*}{No Definition} & small & 7.91 & 7.79 & 6.93 & 7.41 \\
                        & base & 9.91 & 9.79 & 8.98 & 8.51 \\
                        & large & 6.92 & 7.75 & 8.16 & 8.63 \\
                        & 3B & 8.57 & 8.67 & 8.91 & 8.57 \\  \midrule  \midrule
  \multirow{4}{*}{No Ensemble} & small & 13.51 & 17.10 & 37.10 & 51.50 \\
                        & base & 26.40 & 47.10 & 54.40 & 64.59 \\
                        & large & 25.32 & 57.12 & 63.27 & 68.88 \\
                        & 3B & 41.16 & 65.58 & 68.76 & 71.33 \\
    \bottomrule
\end{tabular}
\caption{Performance of TEGEE under different experimental settings, such as Task Definition Generator, LoRA base model, and number of ICL examples. The Human setting is to directly use the task definitions in the dataset. The No Definition setting is to directly train adapter soup without using task definition. The best results in each section are \textbf{Bold}, the second-best
results are \underline{underlined}.}
\label{tab:accents}
\end{table*}

\begin{table*}[htb]
\centering
\resizebox{1\textwidth}{!}{
\begin{tabular}{@{}lcccc|cccc|cccc|cccc@{}}
  \toprule
  Category & \multicolumn{4}{c|}{\makecell{All}} & \multicolumn{4}{|c|}{\makecell{Coreference \\ Resolution}} & \multicolumn{4}{|c|}{\makecell{ Data \\ to Text}} & \multicolumn{4}{|c}{\makecell{ Answerability \\Classification}} \\\midrule
  & Sam 5 & Sam 50 & Sam 100 & Sam 500 & Sam 5 & Sam 50 & Sam 100 & Sam 500 & Sam 5 & Sam 50 & Sam 100 & Sam 500 & Sam 5 & Sam 50 & Sam 100 & Sam 500 \\\midrule
  
        small & 24.95 & 30.17 & 40.66 &52.88 & 23.20 & 30.22 & 42.78 & 53.49 & 16.98 & 20.23 & 28.16 & 34.33 & 30.92 & 38.53 & 58.33 &  67.56 \\\midrule
        base & 37.17 & 50.04 & 56.86 & 64.75 & 32.13 & 52.54 & 60.67 & 65.82 & 26.05 & 34.39 & 36.61 & 40.42 & 55.10 & 66.95 & 68.98 &  77.46 \\\midrule
        large & 40.07 & 59.78 & 65.18 & 69.84 & 39.72 & 65.24 & 69.92 & 76.57 & 30.66 & 38.96 & 40.81 & 43.45 & 62.02 & 73.83 & 76.38 &  81.10 \\\midrule
        3B & 46.81 & 66.57 & 69.55 & 73.52 & 51.78 & 69.30 & 73.93 & 83.61 & 35.93 & 42.80 & 45.10 & 46.07 & 63.79 & 75.99 & 77.84 &  80.77 \\\midrule\midrule

  Category & \multicolumn{4}{c|}{\makecell{All}}  & \multicolumn{4}{|c|}{\makecell{ Question \\ Rewriting}} & \multicolumn{4}{|c|}{\makecell{ Title \\ Generation}} & \multicolumn{4}{|c}{\makecell{ Word \\ Analogy}} \\\midrule

  & Sam 5 & Sam 50 & Sam 100 & Sam 500 & Sam 5 & Sam 50 & Sam 100 & Sam 500 & Sam 5 & Sam 50 & Sam 100 & Sam 500 & Sam 5 & Sam 50 & Sam 100 & Sam 500\\\midrule

        small & 24.95 & 30.17 & 40.66 &52.88 & 47.58 & 51.11 & 57.24 & 60.69 & 15.74 & 19.66 & 23.14 & 30.34 & 8.68 & 9.05 & 18.29 &  58.89 \\\midrule
        base & 37.17 & 50.04 & 56.86 & 64.75 & 55.13 & 59.96 & 61.38 & 63.52 & 22.82 & 32.20 & 33.77 & 36.53 & 12.04 & 36.08 & 69.68 &  99.94 \\\midrule
        large & 40.07 & 59.78 & 65.18 & 69.84 & 50.15 & 62.80 & 65.03 & 65.07 & 26.56 & 35.22 & 37.06 & 38.85 & 17.98 & 80.71 & 97.81 &  100.00 \\\midrule
        3B & 46.81 & 66.57 & 69.55 & 73.52 & 59.70 & 65.09 & 65.66 & 66.34 & 30.42 & 38.33 & 40.33 & 42.06 & 24.10 & 90.54 & 98.20 &  100.00 \\\midrule\midrule

  Category & \multicolumn{4}{c|}{\makecell{All}}  & \multicolumn{4}{|c|}{\makecell{ Textual \\ Entailment}} & \multicolumn{4}{|c|}{\makecell{ Cause Effect \\ Classification}} & \multicolumn{4}{|c}{\makecell{Grammar \\ Error Correction}} \\\midrule
  & Sam 5 & Sam 50 & Sam 100 & Sam 500 & Sam 5 & Sam 50 & Sam 100 & Sam 500 & Sam 5 & Sam 50 & Sam 100 & Sam 500 & Sam 5 & Sam 50 & Sam 100 & Sam 500\\\midrule
        small & 24.95 & 30.17 & 40.66 &52.88 & 22.95 & 30.28 & 43.58 & 59.10 & 35.63 & 41.87 & 52.52 & 53.84 & 62.94 & 65.73 & 71.59 &  78.40 \\\midrule
        base & 37.17 & 50.04 & 56.86 & 64.75 & 43.79 & 56.09 & 63.50 & 71.71 & 43.45 & 54.97 & 56.92 & 65.55 & 73.19 & 80.52 & 81.16 &  83.55 \\\midrule
        large & 40.07 & 59.78 & 65.18 & 69.84 & 42.52 & 64.00 & 72.32 & 80.17 & 49.08 & 60.85 & 65.01 & 74.02 & 69.42 & 82.01 & 84.40 &  83.39 \\\midrule
        3B & 46.81 & 66.57 & 69.55 & 73.52 & 50.28 & 79.75 & 82.54 & 88.04 & 53.63 & 71.44 & 74.67 & 79.04 & 82.99 & 86.21 & 86.58 &  85.96 \\\midrule\midrule
  Category & \multicolumn{4}{c|}{\makecell{All}}  & \multicolumn{4}{|c|}{\makecell{ Overlap \\ Extraction}} & \multicolumn{4}{|c|}{\makecell{ Keyword \\ Tagging}} & \multicolumn{4}{|c}{\makecell{Dialogue \\ Act  Recognition}} \\\midrule  
  & Sam 5 & Sam 50 & Sam 100 & Sam 500 & Sam 5 & Sam 50 & Sam 100 & Sam 500 & Sam 5 & Sam 50 & Sam 100 & Sam 500 & Sam 5 & Sam 50 & Sam 100 & Sam 500\\\midrule
        small & 24.95 & 30.17 & 40.66 &52.88 & 17.62 & 27.70 & 39.24 & 59.07 & 25.55 & 30.71 & 34.41 & 50.39 & 26.18 & 28.77 & 43.71 &  62.83 \\\midrule
        base & 37.17 & 50.04 & 56.86 & 64.75 & 41.76 & 56.40 & 62.40 & 67.86 & 32.08 & 42.72 & 50.78 & 57.49 & 35.69 & 55.12 & 68.81 &  83.51 \\\midrule
        large & 40.07 & 59.78 & 65.18 & 69.84 & 36.44 & 62.71 & 66.27 & 70.07 & 36.03 & 54.18 & 56.51 & 61.19 & 39.33 & 69.61 & 82.26 &  89.62 \\\midrule
        3B & 46.81 & 66.57 & 69.55 & 73.52 & 43.72 & 64.74 & 66.25 & 73.30 & 43.39 & 58.63 & 60.18 & 64.70 & 47.58 & 78.84 & 84.54 &  90.45 \\\bottomrule
        
\end{tabular}
}
\caption{The performance of TEGEE across different task categories.}
\label{tab:categories all results}
\end{table*}

\end{document}